\documentclass[11pt]{article}

\usepackage{eacl2017}
\usepackage{times}
\usepackage{url}

\eaclfinalcopy

\usepackage{url}            
\usepackage{booktabs}       
\usepackage{amsfonts}       
\usepackage{nicefrac}       
\usepackage{microtype}      
\usepackage[acronym]{glossaries}
\usepackage{array,multirow,graphicx}
\usepackage{xspace}
\usepackage[titletoc,title]{appendix}
\usepackage{pbox}
\usepackage{multirow,bigdelim}

\newcommand{\querySeq}{\mathbf{q}}

\newcommand{\documentSeq}{\mathbf{d}}

\newcommand{\answer}{a}
\newcommand{\vocabMat}{\mathbf{V}}

\newcommand{\RUDA}[1]{#1}
\newcommand{\ONDREJ}[1]{#1}

\newacronym[longplural={named entities}]{NE}{NE}{named entity}
\newacronym{CN}{CN}{common noun}
\newacronym{psr}{AS Reader}{Attention Sum Reader}
\newcommand{\asr}{\gls{psr}\xspace}
\newcommand{\bt}{BookTest\xspace}

\title{Embracing data abundance:\\ BookTest Dataset for Reading Comprehension}

\author{
  Ondrej Bajgar\thanks{These authors contributed equally to this work.} , Rudolf Kadlec$^{*}$ \& Jan Kleindienst \\
  IBM Watson\\
  V Parku 4, 140 00 Prague, Czech Republic\\
\texttt{\{obajgar,rudolf\_kadlec,jankle\}@cz.ibm.com} 
}

\begin{document}

\maketitle

\begin{abstract}
There is a practically unlimited amount of natural language data available. Still, recent work in text comprehension has focused on datasets which are small relative to current computing possibilities. This article is making a case for the community to move to larger data and as a step in that direction it is proposing the BookTest, a new dataset similar to the popular Children's Book Test (CBT), however more than 60 times larger.

We show that training on the new data improves the accuracy of our Attention-Sum Reader model on the original CBT test data by a much larger margin than many recent attempts to improve the model architecture. On one version of the dataset our ensemble even exceeds the human baseline provided by Facebook. We then show in our own human study that there is still space for further improvement. 

\end{abstract}

\section{Introduction}

Since humans amass more and more generally available data in the form of unstructured text it would be very useful to teach machines to read and comprehend such data and then use this understanding to answer our questions. A significant amount of research has recently focused on answering one particular kind of questions the answer to which depends on understanding a context document. These are \emph{cloze-style} questions~\cite{taylor1953cloze} which require the reader to fill in a missing word in a sentence. An important advantage of such questions is that they can be generated automatically from a suitable text corpus which allows us to produce a practically unlimited amount of them. That opens the task to notoriously data-hungry deep-learning techniques which now seem to outperform all alternative approaches.

Two such large-scale datasets have recently been proposed by researchers from Google DeepMind and Facebook AI: the CNN/Daily Mail dataset \cite{hermann2015teaching} and the Children's Book Test (CBT) \cite{hill2015goldilocks} respectively. These have attracted a lot of attention from the research community \cite{Kobayashi2016,Kadlec2016,chen2016thorough,Sordoni2016,Dhingra2016,Trischler2016a,Weissenborn2016,Cui2016a,Cui2016,Li2016} with a new state-of-the-art model coming out every few weeks.

However if our goal is a production-level system actually capable of helping humans, we want the model to use all available resources as efficiently as possible. Given that
\begin{enumerate}
\item there is an almost unlimited amount of data that can be used for generating cloze-style question-answering datasets; 
\item it is widely accepted that more data can significantly improve performance of most models~\cite{Banko2001,Halevy2009};
\item training a model takes only about two hours on the CBT or about two days on the Daily Mail dataset, the largest of the three sets, therefore these datasets are not allowing us to fully use current computing potential;
\end{enumerate}
we believe that if the community is striving to bring the performance as far as possible, it should move its work to larger data.


This thinking goes in line with recent developments in the area of language modelling. For a long time models were being compared on several "standard" datasets with publications often presenting minuscule improvements in performance. 
Then the large-scale One Billion Word corpus dataset appeared~\cite{Chelba2014} and it allowed Jozefowicz et al. to train much larger LSTM models~\cite{Jozefowicz2016} that almost halved the state-of-the-art perplexity on this dataset.


We think it is time to make a similar step in the area of text comprehension. Hence we are introducing the \emph{BookTest}, a new dataset very similar to the Children's Book test but more than $60$ times larger\footnote{Therefore more than $16$ times larger than the Daily Mail dataset.}  to enable training larger models even in the domain of text comprehension. Furthermore the methodology used to create our data can later be used to create even larger datasets when the need arises thanks to further technological progress.

We  show that if we evaluate a model trained on the new dataset on the now standard Children's Book Test dataset, we see an improvement in accuracy much larger than other research groups achieved by enhancing the model architecture itself (while still using the original CBT training data). By training on the new dataset, we reduce the prediction error by almost one third. On the named-entity version of CBT this brings the ensemble of our models to the level of human baseline as reported by Facebook~\cite{hill2015goldilocks}. However in the final section we show in our own human study that there is still room for improvement on the CBT beyond the performance of our model.


\section{Task Description}

A natural way of testing a reader's comprehension of a text is to ask her a question the answer to which can be deduced from the text. Hence the task we are trying to solve consists of answering a cloze-style question, the answer to which depends on the understanding of a context document provided with the question. The model is also provided with a set of possible answers from which the correct one is to be selected\footnote{To get open-domain question answering we can provide the whole vocabulary as a candidate list.}. This can be formalized as follows:

The training data consist of tuples $( \querySeq, \documentSeq, \answer, A )$,
where $\querySeq$ is a question, $\documentSeq$ is a document that contains the answer to question $\querySeq$, $A$ is a set of possible answers and $a \in A$ is the ground-truth answer. 
Both $\querySeq$ and $\documentSeq$ are sequences of words from vocabulary $W$\footnote{Note that this vocabulary can be distinct from the model's own dictionary. For instance the model may replace certain rare words by generic unknown-word tags hence reducing its own dictionary size.}. We also assume that all possible answers are words from the vocabulary, that is $A \subseteq W$. In the CBT and CNN/Daily Mail datasets it is also true that the ground-truth answer $a$ appears in the document. This is exploited by many machine learning models~\cite{hill2015goldilocks,Kadlec2016,Sordoni2016,Dhingra2016,Trischler2016a,Cui2016a,Cui2016,Li2016}, however some do not explicitly depend on this property~\cite{hermann2015teaching,Kobayashi2016,chen2016thorough,Weissenborn2016}

\section{Current Landscape}

We will now briefly review what datasets for text comprehension have been published up to date and look at models which have been recently applied to solving the task we have just described.

\subsection{Datasets}

A crucial condition for applying deep-learning techniques is to have a huge amount of data available for training. For question answering this specifically means having a large number of document-question-answer triples available. While there is an unlimited amount of text available, coming up with relevant questions and the corresponding answers can be extremely labour-intensive if done by human annotators. There were efforts to provide such human-generated datasets, e.g. Microsoft's MCTest \cite{Richardson2013}, however their scale is not suitable for deep learning without pre-training on other data~\cite{Trischler2016} (such as using pre-trained word embedding vectors). 

Google DeepMind managed to avoid this scale issue with their way of generating document-question-answer triples automatically, closely followed by Facebook with a similar method. Let us now briefly introduce the two resulting datasets whose properties are summarized in Table~\ref{tab:corpus-stats}.

\begin{table*}
\begin{center}
  \begin{tabular}[t]{lrrrrr}
    \toprule
    & {\bf CNN} & {\bf Daily Mail}& {\bf CBT CN} & {\bf CBT NE} & {\bf BookTest}\\

    \midrule
    \# queries   & 380,298   & 879,450  & 120,769  & 108,719 & 14,140,825\\
    Max \# options & 527     & 371    & 10    & 10  & 10   \\
    Avg \# options & 26.4   & 26.5    & 10    & 10  & 10   \\
    Avg \# tokens  & 762    & 813     & 470   & 433 & 522  \\
    Vocab. size & {118,497} & {208,045} & {53,185} & {53,063} & {1,860,394} \\
    \bottomrule
  \end{tabular}
\end{center}  
  
  \caption{Statistics on the 4 standard text comprehension datasets and our new BookTest dataset introduced in this paper. CBT CN stands for CBT Common Nouns and CBT NE stands for CBT Named Entites. Statistics were taken from \protect\cite{hermann2015teaching} and the statistics provided with the CBT data set.}
    \label{tab:corpus-stats}
\end{table*}

\subsubsection{CNN \& Daily Mail datasets}
These two datasets \cite{hermann2015teaching} exploit a useful feature of online news articles -- many articles include a short summarizing sentence near the top of the page. Since all information in the summary sentence is also presented in the article body, we get a nice cloze-style question about the article contents by removing a word from the short summary.

The dataset's authors also replaced all named entities in the dataset by anonymous tokens which are further shuffled for each new batch. This forces the model to rely solely on information from the context document, not being able to transfer any meaning of the named entities between documents.

This restricts the task to one specific aspect of context-dependent question answering which may be useful however it moves the task further from the real application scenario, where we would like the model to use all information available to answer questions. Furthermore Chen et al.~\cite{chen2016thorough} have suggested that this can make about 17\% of the questions unanswerable even by humans. They also claim that more than a half of the question sentences are mere paraphrases or exact matches of a single sentence from the context document. This raises a question to what extent the dataset can test deeper understanding of the articles.

\subsubsection{Children's Book Test}
The Children's Book Test~\cite{hill2015goldilocks} uses a different source - books freely available thanks to Project Gutenberg\footnote{https://www.gutenberg.org/}. Since no summary is available, each example consists of a context document formed from 20 consecutive sentences from the story together with a question formed from the subsequent sentence. 

The dataset comes in four flavours depending on what type of word is omitted from the question sentence. 
Based on human evaluation done in~\cite{hill2015goldilocks} it seems that \glspl{NE} and \glspl{CN} are more context dependent than the other two types -- prepositions and verbs. Therefore  we (and all of the recent publications) focus only on these two word types.


\subsubsection{Recent additions}
\label{sec:datasets}

Several new datasets related to the (now almost standard) ones above emerged recently. We will now briefly present them and explain how the dataset we are introducing in this article differs from them.

The LAMBADA dataset~\cite{Paperno2016} is designed to measure progress in understanding common-sense questions about short stories that can be easily answered by humans but cannot be answered by current standard machine-learning models (e.g. plain LSTM language models). This dataset is useful for measuring the gap between humans and machine learning algorithms. However, by contrast to our BookTest dataset, it will not allow us to track progress towards the performance of the baseline systems or on examples where machine learning may show super-human performance. Also LAMBADA is just a diagnostic dataset and does not provide ready-to-use question-answering training data, just a plain-text corpus which may moreover include copyrighted books making its use potentially problematic for some purposes. We are providing ready training data consisting of copyright-free books only.

The SQuAD dataset~\cite{Rajpurkar2016} based on Wikipedia and the Who-did-What dataset~\cite{Onishi2016} based on Gigaword news articles are factoid question-answering datasets where a multi-word answer should be extracted from a context document. This is in contrast to the previous datasets, including CNN/DM, CBT, LAMBADA and our new dataset, which require only single-word answers. Both these datasets however provide less than 130,000 training questions, two orders of magnitude less than our dataset does.  

The Story Cloze Test~\cite{Mostafazadeh2016} provides a crowd-sourced corpus of 49,255 commonsense stories for training and 3,744 testing stories with right and wrong endings. Hence the dataset is again rather small. Similarly to LAMBADA, the Story Cloze Test was designed to be easily answerable by humans.


In the WikiReading \cite{Hewlett2016} dataset the context document is formed from a Wikipedia article and the question-answer pair is taken from the corresponding WikiData page. For each entity (e.g. Hillary Clinton), WikiData contain a number of property-value pairs (e.g. place of birth: Chicago) which form the datasets's question-answer pairs. The dataset is certainly relevant to the community, however the questions are of very limited variety with only 20 properties (and hence unique questions) covering $75\%$ of the dataset. Furthermore many of the frequent properties are mentioned at a set spot within the article (e.g. the date of birth is almost always in brackets behind the name of a person) which may make the task easier for machines. We are trying to provide a more varied dataset.

Although there are several datasets related to task we are aiming to solve, they differ sufficiently for our dataset to bring new value to the community. Its biggest advantage is its size which can furthermore be easily upscaled without expensive human annotation. Finally while we are emphasizing the differences, models could certainly benefit from as diverse a collection of datasets as possible.

\subsection{Machine Learning Models}

A first major work applying deep-learning techniques to text comprehension was Hermann et al.~\cite{hermann2015teaching}. This work was followed by the application of Memory Networks to the same task~\cite{hill2015goldilocks}. 
Later three models emerged around the same time~\cite{Kobayashi2016,Kadlec2016,chen2016thorough} including our \asr model~\cite{Kadlec2016}. The AS Reader inspired several subsequent models that use it as a sub-component in a diverse ensemble~\cite{Trischler2016a}; extend it with a hierarchical structure~\cite{Sordoni2016,Shen2016,Dhingra2016}; compute  attention over the context document for every word in the query~\cite{Cui2016a} or use two-way context-query attention mechanism for every word in the context and the query~\cite{Cui2016} that is similar in its spirit to models recently proposed in different domains, e.g.~\cite{Santos2016} in information retrieval.
Other neural approaches to text comprehension are explored in~\cite{Weissenborn2016,Li2016}.



\subsection{Possible Directions for Improvements}

Accuracy in any machine learning tasks can be enhanced either by improving a machine learning model or by using more in-domain training data. Current state of the art models~\cite{Sordoni2016,Dhingra2016,Trischler2016a,Cui2016} improve over AS Reader's accuracy on CBT NE and CN datasets by 1-2 percent absolute. This suggests that with current techniques there is only limited room for improvement on the algorithmic side. 

The other possibility to improve performance is simply to use more training data. The importance of training data was highlighted by the frequently quoted Mercer's statement that ``There is no data like more data.''\footnote{Quote attributed to Robert Mercer by Fred Jelinek \cite{Jelinek2004}} The observation that having more data is often more important than having better algorithms has been frequently stressed since then~\cite{Banko2001,Halevy2009}. 

As a step in the direction of exploiting the potential of more data in the domain of text comprehension, we created a new dataset called \emph{BookTest} similar to, but much larger than the widely used CBT and CNN/DM datasets.


\section{BookTest}

Similarly to the CBT, our BookTest dataset\footnote{BookTest dataset can be downloaded from \url{https://ibm.biz/booktest-v1}. } is derived from books available through project Gutenberg. 
We used $3555$ copyright-free books to extract \gls{CN} examples and $10507$ books for \gls{NE} examples, for comparison the CBT dataset was extracted from just 108 books.


When creating our dataset we follow the same procedure as was used to create the CBT dataset~\cite{hill2015goldilocks}. That is, we detect whether each sentence contains either a named entity or a common noun that already appeared in one of the preceding twenty sentences. This word is then replaced by a gap tag (XXXXX) in this sentence which is hence turned into a cloze-style question. The preceding 20 sentences are used as the context document. For common noun and named entity detection we use the Stanford POS tagger~\cite{Toutanova2003} and Stanford NER~\cite{Finkel2005}\footnote{We use version 3.6.0 of both NER and POS taggers. For POS tagging we use the model stanford-postagger-full-2015-12-09/wsj-0-18-bidirectional-distsim.tagger and for NER we use stanford-ner-2015-12-09/english.all.3class.distsim.crf.ser.gz.}.

The training dataset consists of the original CBT \gls{NE} and \gls{CN} data extended with new \gls{NE} and \gls{CN} examples. The new BookTest dataset hence contains $14,140,825$ training examples and   $7,917,523,807$ tokens.


The validation dataset consists of $10,000$ \gls{NE} and $10,000$ \gls{CN} questions. We have one test set for \glspl{NE} and one for \glspl{CN}, each containing $10,000$ examples. The training, validation and test sets were generated from non-overlapping sets of books.

When generating the dataset we removed all editions of books used to create CBT validation and test sets from our training dataset.
Therefore the models trained on the BookTest corpus can be evaluated on the original CBT data and they can be compared with recent text-comprehension models utilizing this dataset~\cite{hill2015goldilocks,Kadlec2016,chen2016thorough,Sordoni2016,Dhingra2016,Trischler2016a,Weissenborn2016,Cui2016a,Cui2016}.

\section{Baselines}
We will now use our \gls{psr} model to evaluate the performance gain from increasing the dataset size. 

\subsection{AS Reader}

In \cite{Kadlec2016} we introduced the \asr\footnote{\asr is available at \url{https://github.com/rkadlec/asreader}}, which at the time of publication significantly outperformed all other architectures on the CNN, DM and CBT datasets. This model is built to leverage the fact that the answer is a single word from the context document. Similarly to many other models it uses attention over the document -- intuitively a measure of how relevant each word is to answering the question. However while most previous models used this attention as weights to calculate a blended representation of the answer word, we simply sum the attention across all occurrences of each unique words and then simply select the word with the highest sum as the final answer. While simple, this trick seems both to improve accuracy and to speed-up training. It was adopted by many subsequent models~\cite{Trischler2016a,Sordoni2016,Dhingra2016,Cui2016a,Cui2016,Shen2016}.

\begin{figure*}[th]
  \centering
  \includegraphics[width=5in]{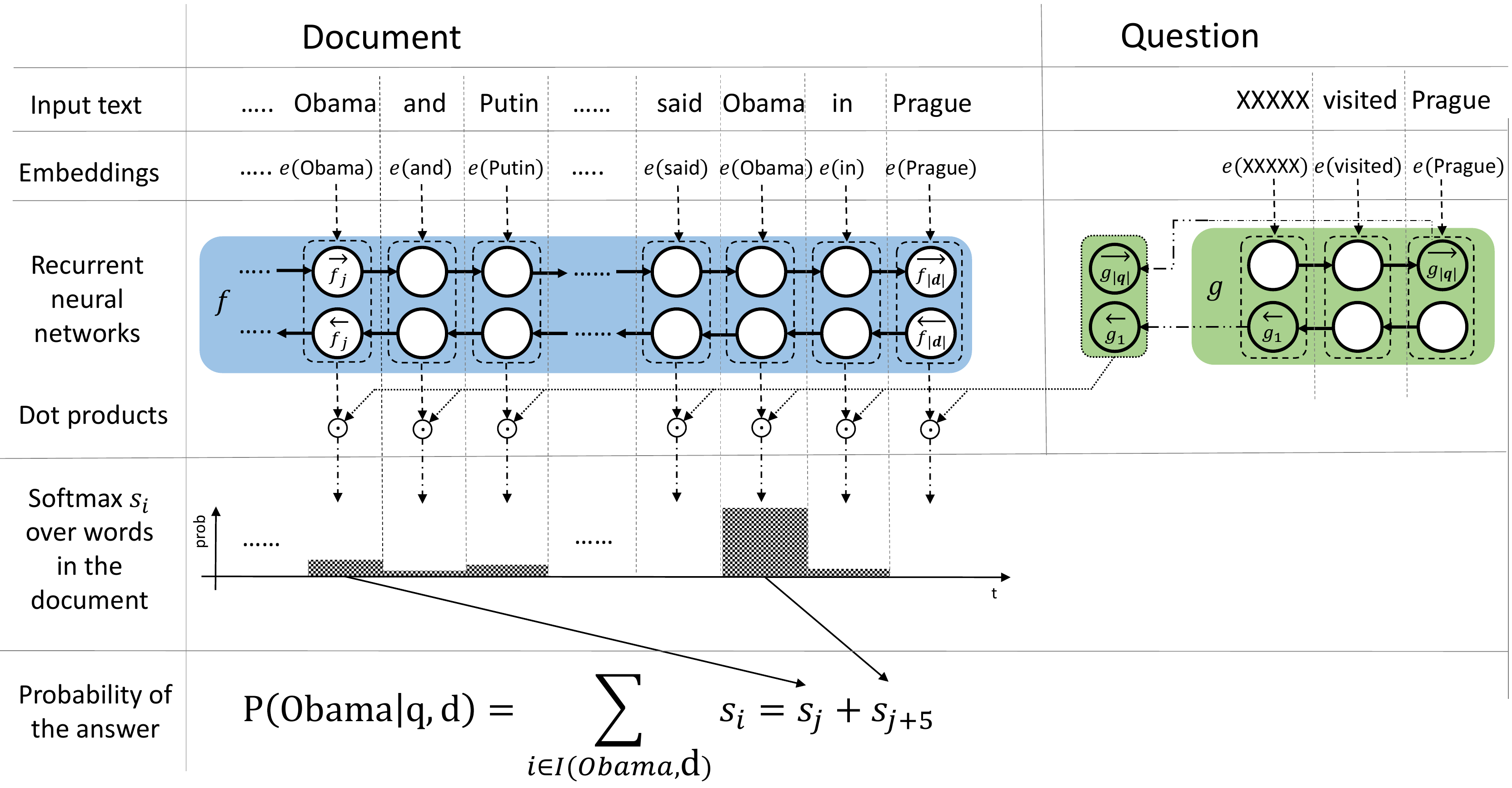}
  \caption   
  {
  Structure of the \asr model. 
  }
  \label{fig:model-structure}
\end{figure*}

Let us now describe the model in more detail. Figure~\ref{fig:model-structure} may help you in understanding the following paragraphs. 

\subsubsection{Basic structure}

The words from the document and the question are first converted into vector embeddings using a look-up matrix $\vocabMat$. The document is then read by a bidirectional GRU network \cite{Cho2014}. A concatenation of the hidden states of the forward and backward GRUs at each word is then used as a \emph{contextual embedding} of this word, intuitively representing the context in which the word is appearing. We can also understand it as representing the set of questions to which this word may be an answer.

Similarly the question is read by a bidirectional GRU but in this case only the final hidden states are concatenated to form the \emph{question embedding}.

The attention over each word in the context is then calculated as the dot product of its contextual embedding with the question embedding. This attention is then normalized by the softmax function and summed across all occurrences of each answer candidate. The candidate with most accumulated attention is selected as the final answer. 

For a more detailed description of the model including equations check~\cite{Kadlec2016}. More details about the training setup and model hyperparameters can be found in the Appendix.

\subsubsection{Out-of-vocabulary words}
During our past experiments on the CNN, DM and CBT datasets \cite{Kadlec2016} each unique word from the training, validation and test datasets had its row in the look-up matrix $\vocabMat$. However as we radically increased the dataset size, this would result in an extremely large number of model parameters so we decided to limit the vocabulary size to $200,000$ most frequent words. For each example, each unique out-of-vocabulary word is now mapped on one of $1000$ anonymous tokens which are randomly initialized and untrained. Fixing the embeddings of these anonymous tags proved to significantly improve the performance.

\subsubsection{Query-initiated encoder}
While mostly using the original AS Reader model, we have also tried introducing a minor tweak in some instances of the model. We tried initializing the context encoder GRU's hidden state by letting the encoder read the question first before proceeding to read the context document. Intuitively this allows the encoder to know in advance what to look for when reading over the context document.

Including models of this kind in the ensemble helped to improve the performance.


\subsection{Results}

\begin{table*}[t]
\centering
  {\begin{tabular}{@{}l@{}r@{}rr@{}r@{}rr@{}c}
    \toprule
    & \multicolumn{2}{c}{Named entity} &\phantom{aa}& \multicolumn{2}{c}{Common noun} & \\
    \cmidrule{2-3} \cmidrule{5-6}
    & valid & test && valid & test  & \\
    \midrule
        Humans (query) \cite{hill2015goldilocks} & NA & 52.0 &&  NA & 64.4 & \\ 
         Humans (context+query) \cite{hill2015goldilocks} & NA & \textbf{81.6} && NA &  \textbf{ 81.6}  & \\ 
        \midrule
         LSTMs (context+query) \cite{hill2015goldilocks} & 51.2 & 41.8 && 62.6 & 56.0 & \rdelim\}{10}{1mm}[\parbox{2cm}{CBT \\training data}]\\
        \midrule
         Memory Networks \cite{hill2015goldilocks}  & 70.4 & 66.6 &&  64.2 &  63.0  & \\ 
        \midrule
         \gls{psr} (single model)     & 73.8 & 68.6 && 68.8 & 63.4  & \\ %
        \textbf{\gls{psr} (avg ensemble)    } & 74.5 &  70.6 && 71.1 & \textbf{68.9}  & \\
        \textbf{\gls{psr} (greedy ensemble)}     &  76.2 &  \textbf{71.0} &&  72.4 &  67.5  & \\
        \midrule
        {GA Reader (ensemble)~\cite{Dhingra2016}}     & {75.5} & {71.9} && {72.1} & {69.4}  & \\ 
        {EpiReader (ensemble)~\cite{Trischler2016a}}    & {76.6} & {71.8} && {73.6} & {70.6}  & \\ %
        \textbf{IA Reader (ensemble)}~\cite{Sordoni2016}    & {76.9} & \textbf{72.0} && {74.1} & \textbf{71.0}  & \\ %
        \textbf{AoA Reader (single model)}~\cite{Cui2016}     & {77.8} & \textbf{72.0} && {72.2} & {69.4}  & \\ %
        
        \midrule
        \midrule
          
        \textbf{\gls{psr} (single model)}     & 80.5 & 76.2 && 83.2 & 80.8 & \rdelim\}{2}{1mm}[\parbox{2cm}{BookTest training data}] \\ 
        \textbf{\gls{psr} (greedy ensemble)}     & 82.3 &  \textbf{78.4} && 85.7 & \textbf{83.7}  & \\

     \bottomrule
  \end{tabular}}
\caption{Results of various architectures on the CBT test datasets.}
\label{tab:results-cbt}
\end{table*}

Table~\ref{tab:results-cbt} shows the accuracy of the \asr and other architectures on the CBT validation and test data. The last two rows show the performance of the \asr trained on the \bt dataset; all the other models have been trained on the original CBT training data.

If we take the best \asr ensemble trained on CBT as a baseline, improving the model architecture as in \cite{Sordoni2016,Dhingra2016,Trischler2016a,Weissenborn2016,Cui2016a,Cui2016}, continuing to use the original CBT training data, lead to improvements of $1\%$ and $2.1\%$ absolute on named entities and common nouns respectively. By contrast, inflating the training dataset provided a boost of $7.4 - 14.8\%$ while using the same model. The ensemble of our models even exceeded the human baseline provided by Facebook \cite{hill2015goldilocks} on the Common Noun dataset.

Our model takes approximately two weeks to converge when trained on the BookTest dataset on a single Nvidia Tesla K40 GPU.

\section{Discussion}
Embracing the abundance of data may mean focusing on other aspects of system design than with smaller data. Here are some of the challenges that we need to face in this situation.




Firstly, since the amount of data is practically unlimited -- we could even generate them on the fly resulting in continuous learning similar to the Never-Ending Language Learning by Carnegie Mellon University~\cite{NELL-aaai15} -- it is now the speed of training that determines how much data the model is able to see. Since more training data significantly help the model performance, focusing on speeding up the algorithm may be more important than ever before. This may for instance influence the decision whether to use regularization such as dropout which does seem to somewhat improve the model performance, however usually at a cost of slowing down training.


Thanks to its simplicity, the \asr seems to be training fast - for example around seven times faster than the models proposed by Chen et al. \cite{chen2016thorough}. Hence the \asr may be particularly suitable for training on large datasets. 

 
 
The second challenge is how to generalize the performance gains from large data to a specific target domain. While there are huge amounts of natural language data in general, it may not be the case in the domain where we may want to ultimately apply our model. 

Hence we are usually not facing a scenario of simply using a larger amount of the same training data, but rather extending training to a related domain of data, hoping that some of what the model learns on the new data will still help it on the original task.

This is highlighted by our observations from applying a model trained on the BookTest to Children's Book Test test data. If we move model training from joint CBT NE+CN training data\footnote{Note that while here we are using joint CBT NE+CN data to create an equivalent of a 230k subset of our BookTest, for most experiments on the CBT we
and some other teams have used NE and CN as two separate training datasests.} to a subset of the BookTest of the same size (230k examples), we see a drop in accuracy of around 10\% on the CBT test datasets.

Hence even though the Children's Book Test and BookTest datasets are almost as close as two disjoint datasets can get, the transfer is still very imperfect
\footnote{This also suggests that the increase in accuracy when using more data that are strictly in the same domain as the original training data results in a performance increase even larger that the one we are reporting on CBT. However the scenario of having to look for additional data elsewhere is more realistic, so we are focusing this article in that direction.}. 
Rightly choosing data to augment the in-domain training data is certainly a problem worth exploring in future work.

Our results show that given enough data the AS Reader was able to exceed the human performance on CBT CN reported by Facebook. However we hypothesized that the system is still not achieving its full potential so we decided to examine the room for improvement in our own small human study.

\section{Human Study}
\label{sec:human-study}

After adding more data we have the performance on the CBT validation and test datasets soaring. However is there still potential for much further growth beyond the results we have observed?

We decided to explore the remaining space for improvement on the CBT by testing humans on a random subset of 50 named entity and 50 common noun validation questions that the \asr ensemble could not answer correctly. 
These questions were answered by 10 non-native English speakers from our research laboratory, each on a disjoint subset of questions.. Participants had unlimited time to answer the questions and were told that these questions were not correctly answered by a machine, providing additional motivation to prove they are better than computers. 
The results of the human study are summarized in Table~\ref{tab:cbt-human-study}. They show that a majority of questions that our system could not answer so far are in fact answerable. This suggests that 1) the original human baselines might have been underestimated, however, it might also be the case that there are some examples that can be answered by machines and not by humans; 2) there is still space for improvement. 

A system that would answer correctly every time when either our ensemble or human answered correctly would achieve accuracy over 92\% percent on both validation and test NE datasets and over 96\% on both CN datasets. Hence it still makes sense to use CBT dataset to study further improvements of text-comprehension systems. 



\begin{table}[t]
\centering{

  \begin{tabular}{l|ccc}
    \toprule
    Dataset & \% correct answers  \\
    
    
    \midrule
    
        Named Entities  & 66\% \\
        Common Nouns  & 82\% \\

     \bottomrule
  \end{tabular}}

\caption{Accuracy of human participants on examples from validation set that were previously incorrectly answered by our system trained only on the BookTest data.}

\label{tab:cbt-human-study}
\end{table}

\section{Conclusion}




Few ways of improving model performance are as solidly established as using more training data. Yet we believe this principle has been somewhat  neglected by recent research in text comprehension. While there is a practically unlimited amount of data available in this field, most research was performed on unnecessarily small datasets.

As a gentle reminder to the community we have shown that simply infusing a model with more data can yield performance improvements of up to $14.8\%$ where several attempts to improve the model architecture on the same training data have given gains of at most $2.1\%$ compared to our best ensemble result. Yes, experiments on small datasets certainly can bring useful insights. However we believe that the community should also embrace the real-world scenario of data abundance.

The \bt dataset we are proposing gives the reading-comprehension community an opportunity to make a step in that direction. 








\bibliographystyle{eacl2017}
\bibliography{booktest.bib}


\begin{appendices}

\begin{table*}[t]
\begin{center}

  \resizebox{0.9\textwidth}{!}{
  \begin{tabular}{l|lll|lll|lll|lll}
    \toprule
     & \multicolumn{3}{c}{Rec. Hid. Layer} & \multicolumn{3}{c}{Embedding} & \multicolumn{3}{c}{Batch size} & \multicolumn{3}{c}{Rec. Layers} \\
    Dataset & min & max & best & min & max & best & min & max & best & min & max & best\\
    
    \midrule
        \bt & 64 & 768 & 384 & 64 & 512 & 128 & 32 & 256 & 128 & 1 & 3 & 2\\ 

     \bottomrule
  \end{tabular}
  }
 \end{center}
  \caption{The best hyperparameters and range of values that we tested on the \bt train dataset. \RUDA{We report number of hidden units of the unidirectional GRU; the bidirectional GRU has twice as many hidden units.}}
  \label{tab:params}

\end{table*}

\section{Training Details}
The training details are similar to those in \cite{Kadlec2016} however we are including them here for completeness.

\RUDA{

To train the model we used stochastic gradient descent with the ADAM update rule~\cite{Kingma2015} and learning rates of $0.0005$, $0.0002$ and $0.0001$. The best learning rate in our experiments was $0.0005$. We minimized negative log-likelihood as the training objective. 

The initial weights in the word-embedding matrix were drawn randomly uniformly from the interval $[-0.1, 0.1]$. Weights in the GRU networks were initialized by random orthogonal matrices~\cite{Saxe2014} and biases were initialized to zero. We also used a gradient clipping~\cite{Pascanu2012} threshold of 10 and batches of sizes between 32 or 256. Increasing the batch from 32 to 128 seems to significantly improve performance on the large dataset - something we did not observe on the original CBT data. Increasing the batch size much above 128 is currently difficult due to memory constraints of the GPU.

During training we randomly shuffled all examples at the beginning of each epoch. To speed up training, we always pre-fetched $10$ batches worth of examples and sorted them according to document length. Hence each batch contained documents of roughly the same length.
}

\RUDA{We also did not use pre-trained word embeddings.}

We did not perform any text pre-processing since the datasets were already tokenized.


During training we evaluated the model performance every 12 hours and at the end of each epoch and stopped training when the error on the 20k BookTest validation set started increasing. 

\RUDA{We explored the hyperparameter space by training 67 different models\footnote{Some of these models were trained on data other than the BookTest which were however very similar.}} \ONDREJ{
The region of the parameter space that we explored together with the parameters of the model with best validation accuracy are summarized in Table~\ref{tab:params}. 
}

Our model was implemented using Theano~\cite{Bastien-Theano-2012} and Blocks~\cite{VanMerrienboer2015}.

The ensembles were formed by simply averaging the predictions from the constituent single models. These single models were selected using the following algorithm.

We started with the best performing model according to validation performance. Then in each step we tried adding the best performing model that had not been previously tried. We kept it in the ensemble if it did improve its validation performance and discarded it otherwise. This way we gradually tried each model once. We call the resulting model a \emph{greedy ensemble}. We used the $20,000$ BookTest validation dataset for this procedure.

The algorithm was offered 10 models and selected 5 of them for the final ensemble. 

\end{appendices}

\end{document}